\documentclass[10pt,twocolumn,letterpaper]{article}

\usepackage{iccv}
\usepackage{times}
\usepackage{epsfig}
\usepackage{graphicx}
\usepackage{amsmath}
\usepackage{amssymb}
\usepackage{multirow}
\usepackage{enumitem}

\usepackage[breaklinks=true,bookmarks=false]{hyperref}

\iccvfinalcopy 

\def\iccvPaperID{7254} 

\ificcvfinal\fi

\begin{document}

\title{Head3D: Complete 3D Head Generation via Tri-plane Feature Distillation}
\author{Yuhao Cheng$^{1}$ \quad 
Yichao Yan$^{1}$\thanks{Corresponding author} \quad 
Wenhan Zhu$^{1}$ \quad 
Ye Pan$^{1}$ \quad 
Bowen Pan$^{2}$ \quad 
Xiaokang Yang$^{1}$ \\ 
$^1$ MoE Key Lab of Artificial Intelligence, AI Institute, Shanghai Jiao Tong University \\
$^2$ Alibaba Group\\
{\tt\small \{chengyuhao, yanyichao, zhuwenhan823, whitneypanye, xkyang\}@sjtu.edu.cn bowen.pbw@alibaba-inc.com}
}

\maketitle
\ificcvfinal\thispagestyle{empty}\fi

\begin{abstract}

Head generation with diverse identities is an important task in computer vision and computer graphics, widely used in multimedia applications. However, current full head generation methods require a large number of 3D scans or multi-view images to train the model, resulting in expensive data acquisition cost. To address this issue, we propose Head3D, a method to generate full 3D heads with limited multi-view images. Specifically, our approach first extracts facial priors represented by tri-planes learned in EG3D, a 3D-aware generative model, and then proposes feature distillation to deliver the 3D frontal faces into complete heads without compromising head integrity. To mitigate the domain gap between the face and head models, we present dual-discriminators to guide the frontal and back head generation, respectively. Our model achieves cost-efficient and diverse complete head generation with photo-realistic renderings and high-quality geometry representations. Extensive experiments demonstrate the effectiveness of our proposed Head3D, both qualitatively and quantitatively.

\end{abstract}

\section{Introduction}

Generating high-fidelity 3D heads poses a significant challenge in the domains of computer vision and graphics, with a broad range of applications, including 3D games and movies. However, existing approaches~\cite{DBLP:conf/iclr/GuL0T22,chan2022efficient,DBLP:conf/cvpr/XuPYSZ22,DBLP:conf/cvpr/Or-ElLSSPK22,DBLP:conf/cvpr/XueLSL22,DBLP:conf/cvpr/ZhangZGZPY22,DBLP:journals/corr/abs-2110-09788,ploumpis2019combining} primarily concentrate on frontal faces, lacking the capacity of rendering the side or back views, thus their applications are significantly limited. In this work, we aim to address the issue and generate complete 3D heads with photo-realistic rendering capabilities.


\begin{figure}[t]
\centering
    	\begin{minipage}[h]{1.0\linewidth}
    	\centering
    	\includegraphics[width=1.0\linewidth]{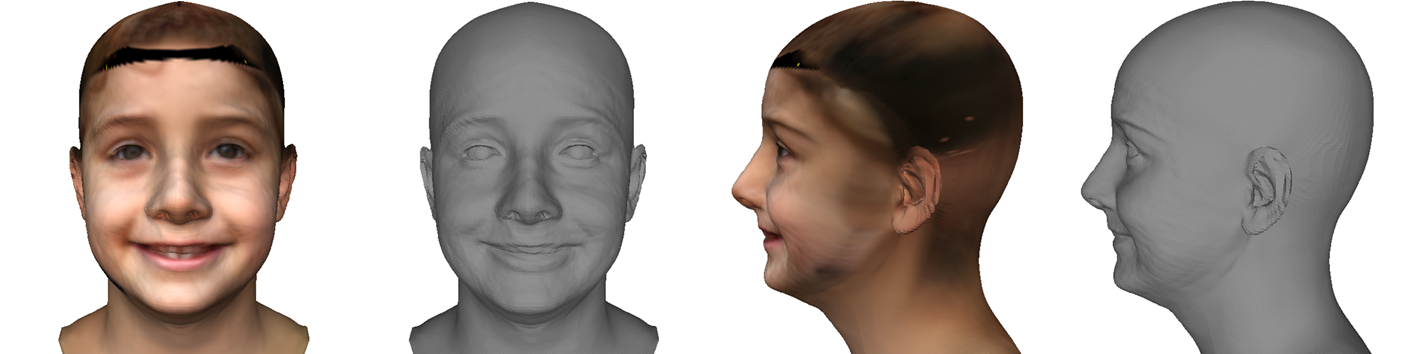}
    	\small (a)~Example of parametric models, DECA~\cite{feng2021learning}
    	\includegraphics[width=1.0\linewidth]{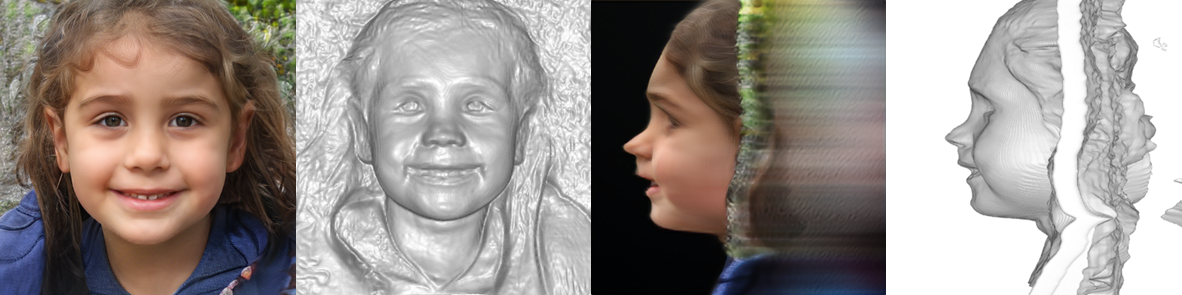}
    	\small (b)~Example of 3D-aware GAN trained by 2D data, EG3D~\cite{chan2022efficient}
    	\includegraphics[width=1.0\linewidth]{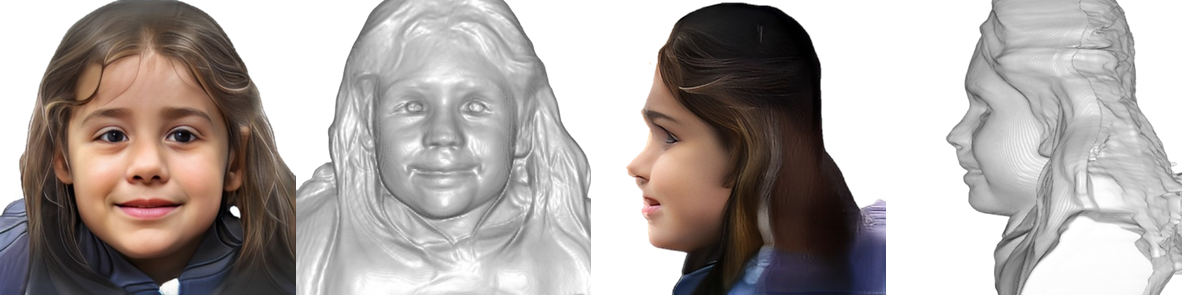}
    	\small (c)~Visualization of Our Head3D
	    \end{minipage}%
	\centering
	\caption{The illustration of existing head/face generation methods and our presented Head3D. All results are rendered into RGB images and geometric form, in both frontal and side views.}
\vspace{-2em}
\label{img:beginning}
\end{figure}


Current 3D head generation methods fall into two categories: non-parametric head models and parametric head models. Non-parametric methods~\cite{ramon2021h3d,Grassal_2022_CVPR,wang2022morf,burkov2022multi} predict 3D heads from single view or multi-view images, usually considered as \textbf{``reconstruction''} methods. These methods face challenge in reproducing heads that do not exist, limiting the variety of generated results. Moreover, limited-view reconstruction results in insufficient details due to the absence of visible perspectives.
Parametric models~\cite{booth20163d,paysan20093d,booth20173d,gerig2018morphable,li2017learning,tran2019towards,cao2013facewarehouse,patel20093d,ploumpis2020towards,dai20173d,yenamandra2021i3dmm,giebenhain2022learning}~(Figure~\ref{img:beginning} (a)) utilize decoupled parameters to represent heads, which rely on a vast of expensive 3D scans and are hard to express intricate texture and geometry. 
Learning 3D head generation solely from images can be a more cost-effective approach to address the challenging task, and it has the potential to generate richer identities and higher-fidelity outcomes.
Recently 3D-aware GANs~\cite{DBLP:conf/iclr/GuL0T22,DBLP:conf/cvpr/Or-ElLSSPK22,chan2021pi,chan2022efficient} are learned from easily accessible in-the-wild images to generate 3D frontal faces with photo-realistic rendering and high-fidelity shapes~(Figure~\ref{img:beginning} (b)). These methods can also be used in generating heads theoretically. However, accurate camera poses are crucial for 3D consistency in these methods~\cite{chan2022efficient}, while estimating them without landmarks on the back is challenging. Hence, our objective is to address the aforementioned challenges and devise an approach capable of generating complete heads solely by training on limited images.

Motivated by the high-fidelity 3D face generator~\cite{chan2022efficient}, a question arises: can we use it as prior knowledge to generate full heads? We answer this question with \emph{YES}, but two challenges must be addressed. First, \emph{how to extract the 3D priors of heads?} The face prior is represented implicitly, making it difficult to integrate the face topology directly with the head in a re-topological manner akin to computer graphics. A straightforward idea is to directly fine-tune the 3D-aware generation model on full head data. However, fine-tuning the pre-trained model with limited data, \eg, several thousand images, often leads to mode collapse or over-fitting~\cite{gal2022stylegan}, resulting in limited face diversity and low quality. Second, \emph{how to bridge the domain gap between the frontal faces and the hair?} 
Obviously, the frontal face and back of the haired heads share two related but different distributions, respectively. 
Moreover, obtaining back-view images with accurate view direction is challenging, resulting in highly imbalanced quantities in these two distributions. This poses an extra requirement for the discriminator in the 3D-aware GAN model to not only distinguish real/fake samples, but also frontal/back views.


In this work, we propose Head3D for diverse full head generation that builds on a current SOTA 3D face generative model, \ie, EG3D~\cite{chan2022efficient}. Our goal is to transfer the facial prior generated by EG3D onto full heads with limited 3D data. 
To extract the 3D prior, we conduct a systematic analysis of the tri-plane representation and observe a decoupling between the geometric and identity information in this representation. Based on this observation, we propose a tri-plane feature distillation framework that aims to preserve the identity information while completing the head geometry.
To address the second problem, we design dual-discriminators for frontal faces and the back of heads respectively, which not only frees the discriminator from distinguishing frontal/back images and also inherits the strong capability of discriminators from EG3D.


To evaluate the effectiveness of our Head3D, we conduct comparative analyses with the original EG3D~\cite{chan2022efficient} and other recent baselines. Our experiments demonstrate that the proposed Head3D model can produce superior results compared to previous approaches, despite being trained with only a small amount of multi-view images and prior knowledge. Examples of our model are shown in Figure~\ref{img:beginning} (c). Additionally, we perform ablation studies to validate the effectiveness of each component in Head3D.

The main contributions are summarized as follows: 
\textbf{(a)} We propose the novel Head3D for generating full heads with rich identity information, photo-realistic renderings, and detailed shapes with limited 3D data. \textbf{(b)} We investigate the effectiveness of tri-plane features, and propose tri-plane feature distillation to enable the transfer of identity information onto the head templates. \textbf{(c)} We propose a dual-discriminator approach to address the distribution gap and quantity imbalance between front-view and back-view images, thereby ensuring the quality of generated heads.

\section{Related Work}

\subsection{Head Generation Model}
As previously discussed in introduction, head generation can be categorized into two types. \textbf{``Reconstruction''} methods\cite{ramon2021h3d,wang2022morf,Grassal_2022_CVPR,burkov2022multi} primarily learn the correspondence between 3D data and images to establish prior knowledge. When presented with new images, these methods optimize the difference between the reconstructed results and the images to accomplish reconstruction. However, these methods are limited to the reconstruction of heads present in the given images, and are incapable of generating novel heads.
Explicit parametric 3D morphable models~\cite{blanz1999morphable,booth20163d,paysan20093d,booth20173d,gerig2018morphable,li2017learning,tran2019towards,cao2013facewarehouse,patel20093d,ploumpis2020towards,dai20173d,ploumpis2019combining} represent identities, textures and expressions by low dimensional PCA parameters, which are learned from 3D scans with different expressions and identities. Similarly, implicit 3D parametric models~\cite{yenamandra2021i3dmm,giebenhain2022learning,hong2022headnerf} typically employ an auto-decoder to learn the decoupled parameters from 3D scans. However, these methods are trained with large number of expensive 3D scans and are hard to express detailed texture and geometry. 
Recent Rodin~\cite{wang2022rodin} employs a diffusion model to learn head generation trained by images. However, this approach requires a large dataset consisting of 100,000 3D models for rendering multi-view images, and each identity is reconstructed with tri-plane alone for training the diffusion model. The employed dataset and training procedure are particularly expensive.
In contrast to these expensive methods, our aim is to generate diverse novel and high-fidelity heads in a cost-effective manner, utilizing only implicit face priors and a limited amount of multi-view images.


\subsection{NeRF-based GAN Model}
Neural radiance field (NeRF)~\cite{mildenhall2021nerf} represents 3D models by implicit networks, whose outputs are density and color under the input of position and view direction, optimized by view reconstruction between volume rendering results and ground truths. Recent methods, integrating NeRF and GANs~\cite{goodfellow2020generative}, aim at learning 3D-aware generators from a set of unconstrained images~\cite{schwarz2020graf,chan2021pi,genova2020local,meng2021gnerf,niemeyer2021giraffe,sun2022fenerf,DBLP:journals/corr/abs-2206-10535}. GRAF~\cite{schwarz2020graf} verifies NeRF-based GAN can generate 3D consistent model with high fidelity rendering results. Pi-GAN~\cite{chan2021pi} introduces SIREN~\cite{sitzmann2020implicit} and a growing strategy for higher-quality image synthesis results. To improve rendering efficiency, several works volumetrically render a low-resolution feature, then up-sample them for high-resolution view synthesis under different 3D consistency constraints~\cite{DBLP:conf/iclr/GuL0T22,chan2022efficient,DBLP:conf/cvpr/XuPYSZ22,DBLP:conf/cvpr/Or-ElLSSPK22,DBLP:conf/cvpr/XueLSL22,DBLP:conf/cvpr/ZhangZGZPY22,DBLP:journals/corr/abs-2110-09788}. Particularly, EG3D~\cite{chan2022efficient} provides a hybrid 3D representation method, which first generates a tri-plane features, then sampled features are decoded and rendered for image synthesis. Pose-related discrimination is utilized for 3D consistency generation. Surrounding head images are prerequisite for these methods to achieve complete heads generation. 
However, it is difficult to predict the view directions of in-the-wild back head images, whereas pose-accurate multi-view images are expensive. Our Head3D extends EG3D to achieve complete head generation with limited multi-view head images by facial priors transfer via tri-plane feature distillation.


\subsection{Knowledge Distillation of GANs}
The target of knowledge distillation(KD)~\cite{DBLP:journals/corr/HintonVD15} is to transfer dark knowledge, \eg logits and features, from teacher networks to student network, which is originally used in classification~\cite{Cho2019efficacy, park2019relational,romero2014fitnets,DBLP:conf/eccv/XuLLL20,DBLP:conf/iclr/TianKI20,zhang2020prime}. Following, KD is also employed for GAN-based model compression~\cite{chen2020distilling,wang2020gan,li2020gan,liu2021content,xu2022mind,hou2021slimmable}. 
In GAN compression~\cite{li2020gan}, the student network learns every intermediate features and final outputs from the teacher network. Besides conditional GANs, several works focus on the study of unconditional GANs. CAGAN~\cite{liu2021content} adopts multi-Level distillation and content-aware distillation, then fine-tuned by adversarial loss. StyleKD~\cite{xu2022mind} proposes that the mapping network plays an important role in generation. In addition, a novel initialization strategy and a latent direction-based distillation loss are presented for semantic consistency between the teacher and student model. 
Our work is built upon a StyleGAN-like 3D-aware generative model EG3D~\cite{chan2022efficient}. Through experiments, we observe the disentanglement of tri-planes, which are critical in semantic representations in EG3D. Accordingly, a tri-plane feature distillation procedure is proposed to transfer facial priors in complete head generation.

\section{Methods}
With a small number of multi-view head images, our goal is to achieve complete heads generation with pre-trained face generator. We first review an efficient and effective tri-plane-based face generator EG3D (Sec~\ref{preliminary}). Then, aiming to extract the semantic information of the face as a prior knowledge, we explore how tri-planes represent the faces (Sec~\ref{Findings}). Afterward, we describe how to apply the prior knowledge to generate full heads (Sec~\ref{pipeline}). Finally, the training process is introduced in detail (Sec~\ref{training}). 

\subsection{Revisit Tri-planes for 3D Generation}
\label{preliminary}
EG3D is a tri-plane-based generative model. Similar to StyleGAN~\cite{karras2019style,Karras_2020_CVPR}, mapping networks $M$ process the input latent code $z$ and camera params $p$ to style code $w$:
\begin{equation}
	w = M(z, p).
\end{equation}

Then, feature maps $F$ are extracted via a CNN-based generator $C$, yielding three planes which are rearranged orthogonally to form a tri-plane structure:
\begin{equation}
	F_{xy}, F_{xz}, F_{yz} = C(w),
\end{equation}
where $F_{xy}$, $F_{xz}$ and $F_{yz}$ are three planes in the front, side and top views, respectively. The tri-plane features contain semantic information of faces, determining their identities.

\begin{figure}[t]
	\centering
	\includegraphics[width=1.0\linewidth]{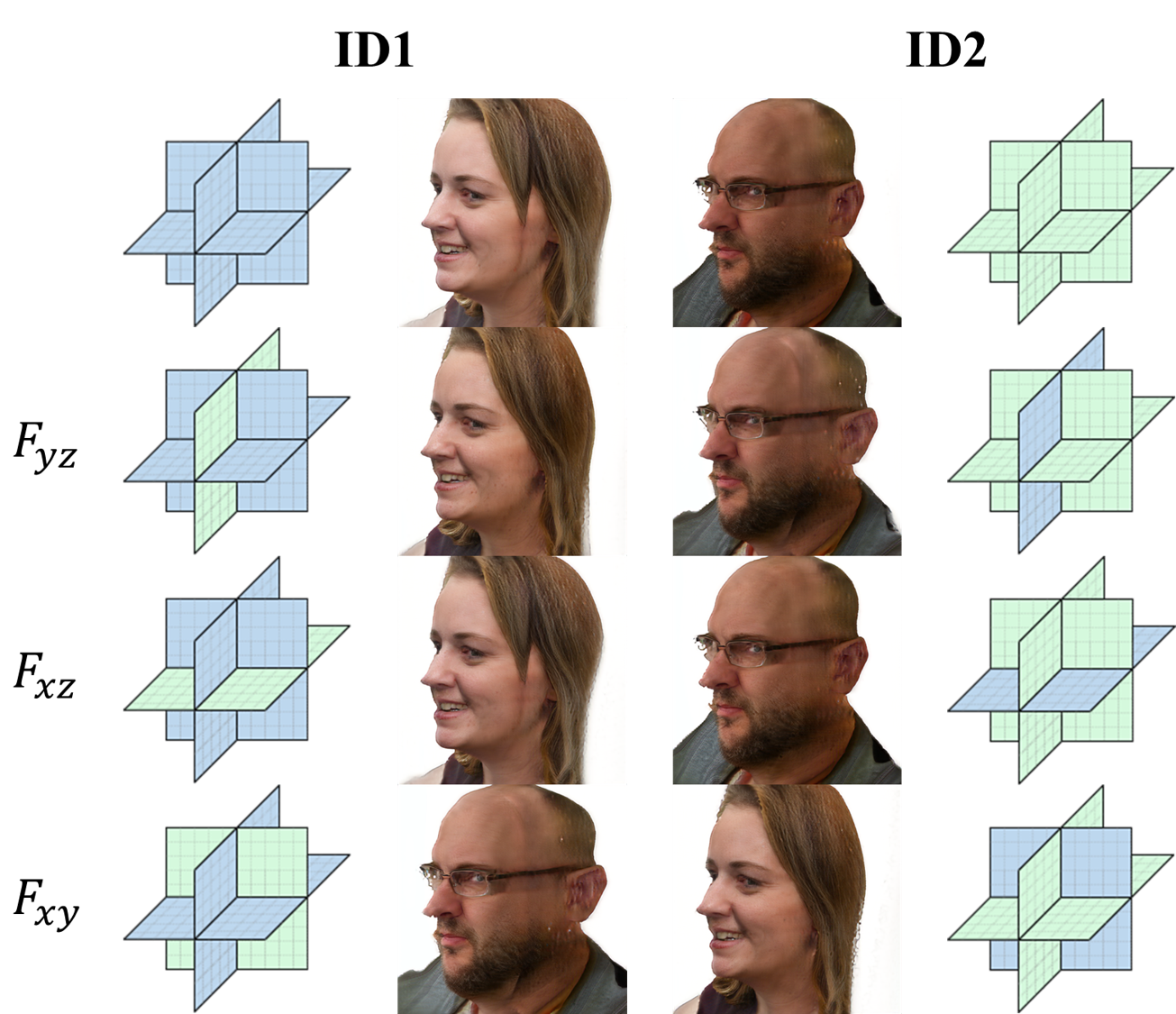}
	\caption{Visualization analysis of exchanging each plane of tri-plane. ID 1 and ID 2 are generated from the same latent code $z$ and camera parameter $p$.}
	\label{fig:short-a}
\end{figure}

\begin{table}[t]
\small
	\centering
	\setlength{\tabcolsep}{5mm}{\begin{tabular}{ccccc}
			\hline
			$F_{xy}$ & $F_{xz}$ & $F_{yz}$ & ID 1 & ID 2  \\
			\hline
			ID 2 & ID 1 & ID 1 &0.535 & 0.968\\
			ID 1 & ID 2 & ID 1 &0.977 & 0.534\\
			ID 1 & ID 1 & ID 2 &0.988 & 0.532\\
			ID 1 & ID 1 & ID 1 &1.000 & 0.531\\
			ID 2 & ID 2 & ID 2 &0.531 & 1.000 \\
			\hline
	\end{tabular}}
	\caption{Quantitative results with tri-plane exchanged. ID 1 and ID 2 denote two different identities generated from different $z$ by EG3D. We report the average results between every pairs between 100 identities.} 	
	\label{lab:observationa}
	\vspace{-1em}
\end{table}

Afterwards, features of locations are sampled from tri-planes, and are fed into decoder to output densities $\sigma$ and colors $c$. Then, volume rendering is performed to obtain a moderate resolution images $I_{raw}$.
\begin{equation}
I_{raw}=\int_0^{\infty} p(t) \boldsymbol{c}(\boldsymbol{r}(t), \boldsymbol{d}) d t, 
\end{equation}
where $p(t)=
\exp \left(-\int_0^t \sigma(\boldsymbol{r}(s)) d s\right) \cdot \sigma(\boldsymbol{r}(t))$, $\boldsymbol{r}(t)$ represents camera ray, and $t$ is the distance from camera.

Finally, a super-resolution module $S(\cdot)$ are performed to up-sample the $I_{raw}$ to results $I$ of high resolution:
\begin{equation}
\begin{aligned}
 I = S(I_{raw}).
\end{aligned}
\end{equation}

\begin{figure}[t]
	\centering
	\includegraphics[width=1.0\linewidth]{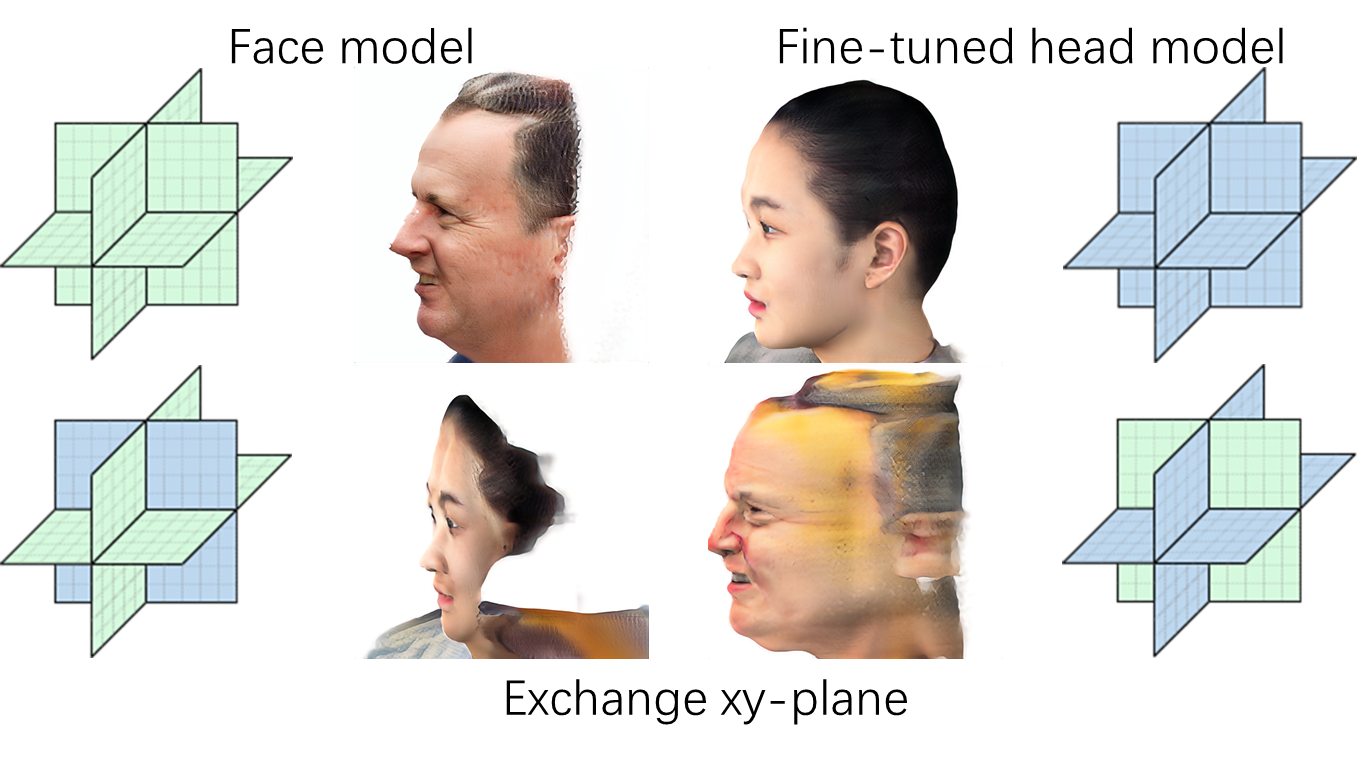}
	\caption{Visualization analysis of exchanging $xy$-plane between face generator and head generator.}
	\label{fig:assumpt}
 \vspace{-1em}
\end{figure}

\subsection{Dientangled Representation of Tri-planes}
\label{Findings}
Insufficient multi-view head images pose a challenge for generating diverse complete heads. However, leveraging the existing EG3D model, which can generate high-quality 3D frontal faces, offers a potential solution. Similar to StyleGAN, the semantic results in EG3D are determined by the style codes $w$, which generate tri-plane features as the only output of the CNN generator $C$. As a result, all the semantic information is contained within tri-planes. To enable knowledge transfer, we conduct an in-depth analysis of tri-planes to explore how prior knowledge is represented.

To investigate the role of each plane, we exchange each plane ($F_{xy}$, $F_{xz}$ and $F_{yz}$) between two different samples, and then generate faces from the newly integrated tri-plenes for visual analysis. The results are shown in Figure~\ref{fig:short-a}. We observe that the identities are exchanged along with $F_{xy}$ changed, while the identities remain the same after exchanging $F_{yz}$ and $F_{xz}$. Additionally, we conduct numerical experiments on 100 different identities. We exchange $F_{xy}$, $F_{xz}$ and $F_{yz}$ for each identify pair, respectively, and render the tri-planes to images from the frontal view. After that, the identity consistency between the exchanged tri-plane and the original identities are measured by Arcface~\cite{deng2019arcface}, and the results are calculated by averaging all the sample pairs. Table~\ref{lab:observationa} indicates that the identities are preserved when $F_{yz}$ and $F_{xz}$ changed, whereas they exchange when $F_{xy}$ are exchanged. Based on these observations, we preliminarily assume that $F_{xy}$ plays a crucial role in determining the identity information.

To further validate our assumption, we fine-tune a head generator from EG3D and perform experiments where $F_{xy}$ is exchanged between faces and heads. 
The instanced results, as shown in Figure~\ref{fig:assumpt}, indicate that exchanging $F_{xy}$ leads to identity transfer while do not influence the geometry whether to be heads or faces.
Our findings support the notion that $F_{xy}$ primarily controls the identity information, while the other two planes, $F_{yz}$ and $F_{xz}$, mainly represent the geometric shape of the head. These results provide evidence for our proposed decoupling between identity and geometric information in the tri-plane representation.

This observation is consistent with our expectation, as the $xy$-plane is aligned with the training images to better capture the facial features, while the other two planes are orthogonal to the frontal view, which mainly represent the depth and geometric information.
The experimental results and analysis sheds some light on the effect of tri-plane features in EG3D and offer a starting point for considering how to employ this prior for complete head generation.

\begin{figure*}[htpb]
	\centering
	\includegraphics[width=.85\textwidth]{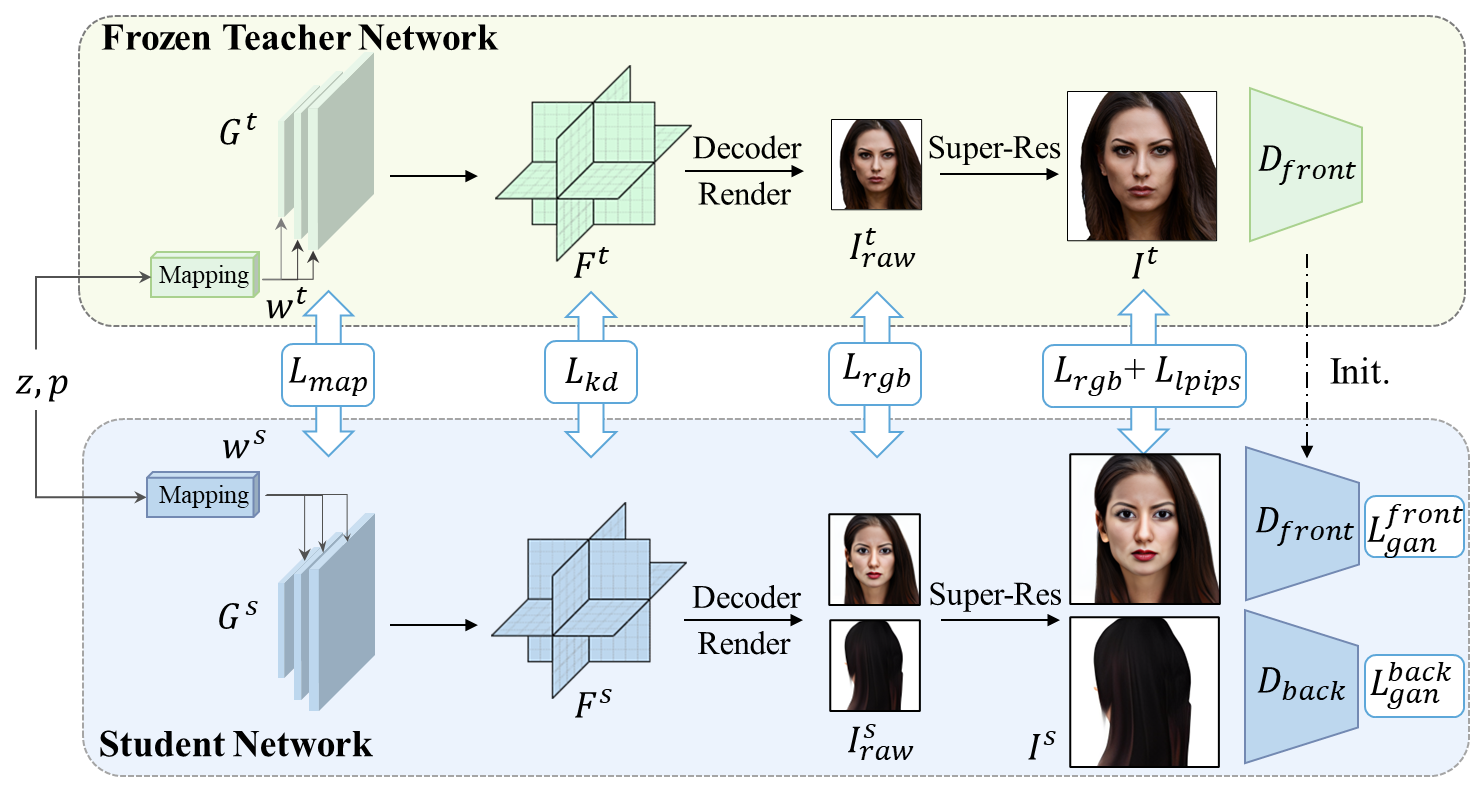}
	\caption{The overview of our proposed Head3D. First, the frozen pre-trained EG3D is served as the teacher network $G^t$, while the fine-tuned head generator is student network $G^s$. Then tri-plane feature distillation and multi-level loss functions are employed for photo-realistic and diversy full head generator. Note that, the parameter $p$ for rendering and discriminator are not displayed in this framework.}
	\label{img:overview}
	\vspace{-1em}
\end{figure*}

\subsection{Tri-plane Distillation for 3D Head Generation}
\label{pipeline}
One potential approach to generate complete heads with the frontal face priors is to directly fine-tune the face generator with a few multi-view data. However, training with a small number of data can cause mode collapse or over-fitting, resulting in limited diversity and low quality. 
Inspired by computer graphic methods where the face is extracted from a photo and then registered with the head template model, we can also apply the tri-plane-based face priors to an implicit head template. 
Additionally, knowledge distillation~\cite{DBLP:journals/corr/HintonVD15} is a general method to transfer dark knowledge between models, allowing for the delivery of facial priors. Therefore, we propose a tri-plane feature distillation method, as illustrated in Figure~\ref{img:overview}. First, we fine-tune a full head generator from the pre-trained EG3D using a small amount of multi-view head images. Then, served as prior knowledge, $F_{xy}$ are transferred from the face generator to the head network via knowledge distillation to ensure consistency in identity. Benefiting from the powerful presentation capability of tri-planes, we are able to integrate the head template and face priors, enabling diverse full head generation.






In order to achieve photo-realistic rendering results, GAN loss is necessary in training the tri-plane distillation framework~\cite{xu2022mind}. 
The importance of camera poses in learning correct 3D priors in the discriminator has been highlighted in EG3D~\cite{chan2022efficient}.
For human heads, camera poses can be categorized into front and back perspectives to guide the generation of the face and back, respectively.
A large number of single-view face datasets are available to provide front images, whereas the number of back images is limited to half of the small multi-view dataset. As a result, there is an imbalance in the quantity and a distribution gap in perception between frontal and back head images. It is hard for a single discriminator to simultaneously guide fine-grained face generation and guarantee full head geometry in two different domains. Referring to this, we propose a dual-discriminator to ensure the generation quality and maintain the head completed. In our method, two discriminators guide the generation of front and back images, respectively. Moreover, two discriminators are alternated during training to mitigate the effects of imbalanced data.

\subsection{Model Training}
\label{training}

\noindent \textbf{Face Prior Transfer.}
As depicted in Figure~\ref{img:overview}, we present Head3D, a tri-plane distillation approach for generating diverse heads. Firstly, a head generator $G_s$ is fine-tuned from a pre-trained face generator $G_t$ with scarce multi-view head images. Then, to ensure identity consistency, the transformation of $F_{xy}$ is calculated by L2-norm, represented as:
\begin{equation}
	\begin{aligned}
		L_{kd} = {||F^t_{xy} - F^s_{xy}||}_2.
	\end{aligned}
\end{equation}
where $F^t_{xy}$ and $F^s_{xy}$ are $xy$-plane features generated by head generator $G_t$ and face generator $G_s$, respectively.

 \begin{figure*}[t]
	\centering
	\includegraphics[width=1.0\linewidth]{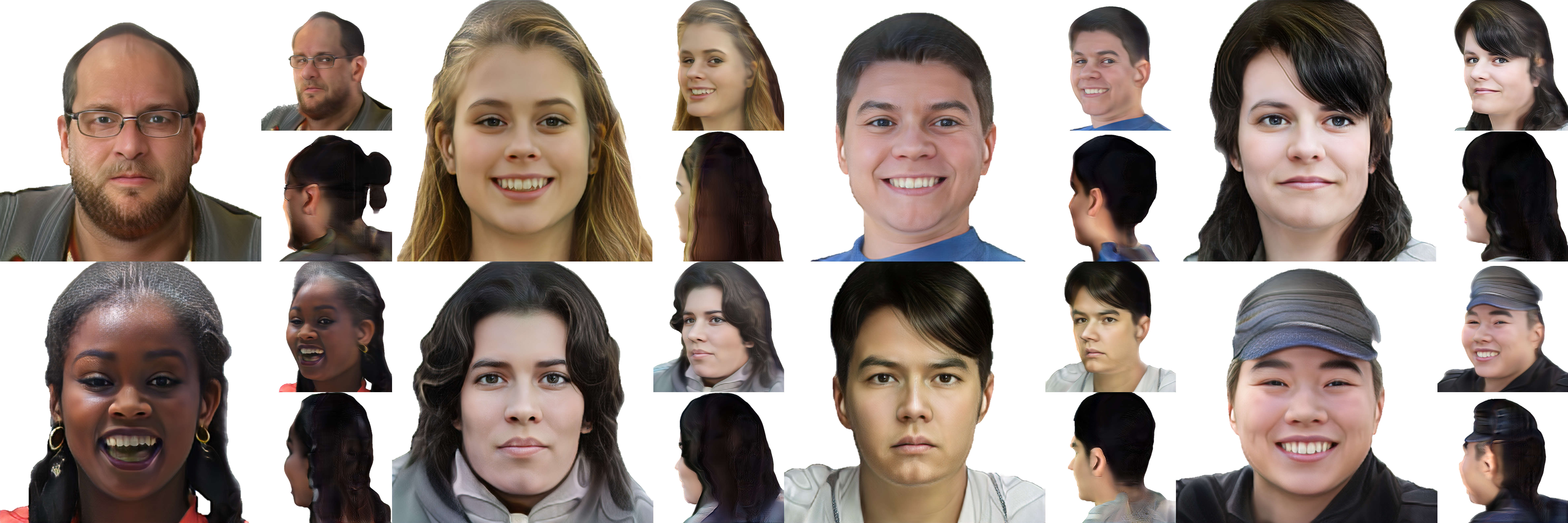}
	\caption{Example results of our proposed Head3D, synthesized from three different views.}
	\label{img:show}
 \vspace{-1.5em}
\end{figure*}

\noindent \textbf{Dealing with Distribution Gap.}
To realize high-fidelity front and back head generation, we employ two discriminators $D_{front}$ and $D_{back}$ for training, which share the same network structure as the discriminator in EG3D.
The purpose of $D_{front}$ is to ensure the quality of faces, and it is initialized by the pre-trained face EG3D. $D_{back}$ is to ensure the integrity of the head, which is initialized by the fine-tuned head generator. Fed together with camera parameters $p$, both discriminators support the synthesis results $I$ be 3D consistent and in a similar distribution with ground truth images $I_{gt}$. The GAN loss function can be represented as:
\begin{equation}
	\begin{aligned}
			L_{gan} &	=
		\mathbb{E}[f(D(p, \mathbf{I})) 
		+  f\left(-D\left(p, \mathbf{I}_{gt}\right)\right) ] \\
		&+ \gamma{||\nabla D\left(p, \mathbf{I}_{gt}\right) ||}^2,
	\end{aligned}
\end{equation}
where $f(x)=-{\rm log}(1+{\rm exp}(-x))$, and $\gamma$ is a hyper-parameter in R1 regularization. Dual-discriminators are trained separately according to the view from face or back of heads, whose loss functions are represented as $L^{front}_{gan}$ and $L^{back}_{gan}$, respectively. In addition, the regularization coefficient $\gamma$ of dual-discriminator is different due to the imbalanced quantity of dataset.

\noindent \textbf{Detailed Texture Learning.}
StyleKD~\cite{xu2022mind} proposes that mapping network determines semantic informations of generators. Therefore, besides tri-plane feature distillation, it is also necessary to ensure that the output $w$ of the mapping network is consistent. Following StyleKD~\cite{xu2022mind}, a mapping loss is utilized:
\begin{equation}
	\begin{aligned}
		L_{map} = {||W^t - W^s||}_1.
	\end{aligned}
\end{equation}

Moreover, in order to learn more detailed faces, RGB loss and LPIPS loss~\cite{zhang2018unreasonable} are applied only to frontal neural renderings $I_{raw}$ and super-resolutioned results $I$. Referring to the quality of the generated results of original EG3D declines in the side view, this loss function is applied only to the rendering results within a certain range.
\begin{equation}
	\begin{aligned}
		L_{rgb} = \mathbb{I}(|\Delta p|\leq \tau)[{||I^t - I^s||}_1 + {||I^t_{raw} - I^s_{raw}||}_1],
	\end{aligned}
\end{equation}
\begin{equation}
	\begin{aligned}
		L_{lpips} &= \mathbb{I}(|\Delta p| \leq \tau){||F(I^t) - F(I^s)||}_1, 
	\end{aligned}
\end{equation}
where $F(\cdot)$ is a well-trained frozen VGG~\cite{DBLP:journals/corr/SimonyanZ14a} to extract multi-scale semantic information from images. $\Delta p$ is the horizontal offset angle from center, and $\tau$ is a threshold. These two loss functions work in image space and perceptual space respectively, ensuring detailed consistency between the teacher and student.

Finally, the final loss function is weighted sum with the above loss functions:
\begin{equation}
	\begin{aligned}
		L &=	\lambda_{gan_{front}}L^{front}_{gan} + \lambda_{kd}L_{kd} + \lambda_{rgb}L_{rgb} \\
		&+\lambda_{lpips}L_{lpips}+\lambda_{map}L_{map}  
		+ \lambda_{gan_{back}}L^{back}_{gan},
	\end{aligned}
\end{equation}
where $\lambda_{*}$ denotes the weights of each loss functions.


\section{Experiments}

\subsection{Experiments setup}

\noindent\textbf{Dataset.}
We train our proposed Head3D using two datasets. One is FFHQ~\cite{karras2019style}, a large public single-view real-world face dataset, for face priors learning. And another is H3DH, our proposed multi-view human head dataset. H3DH contains multi-view images of 50 identities, who are gender- and age-balanced with a wide variety of hairstyles. Following EG3D, we employ the same off-the-shelf pose estimators~\cite{deng2019accurate} to extract approximating camera extrinsic of FFHQ. In terms of H3DH, we set the same camera intrinsic as that of FFHQ, and obtain images with the resolution of $512^2$ from surrounding perspectives under natural light.


\noindent\textbf{Implementation Details.}
We fine-tune the head generator in the proceducre from StyleGAN-ADA\cite{karras2020training} via the H3DH dataset with batch size 16. The optimizer is Adam~\cite{DBLP:journals/corr/KingmaB14} with the same learning rate as original EG3D~\cite{chan2022efficient} of 0.0025 for G and 0.002 for D. D is trained with R1 regularization with $\gamma=20$. In the knowledge distillation phase, the learning rate is converted to 0.001 for G and 0.0005 for D, with $\gamma=1$ for $D_{front}$ and $\gamma=20$ for $D_{back}$. All $\lambda_{*}$ are set to 1.0 except $\lambda_{kd}$ set to 0.5 and $\lambda_{gan_{back}}$ set to 10.0 to balance loss functions in the front view and back view. Threshold $\tau$ is set to $\pi/4$. The resolutions of neural rendering $I_{raw}$ and final generated images $I$ are ${128}^2$ and ${512}^2$, respectively. Note that, owing to H3DH rendered without background, we regard fine-tuned EG3D which generates faces in the white background as the teacher network instead of the original EG3D, which is fine-tuned with images in FFHQ whose background is removed by BiseNet~\cite{yu2018bisenet}. Training a model costs about 4 hours on two Nvidia A100 GPUs. 

\subsection{Comparisons}
 \begin{figure}[t]
	\centering
	\includegraphics[width=1.0\linewidth]{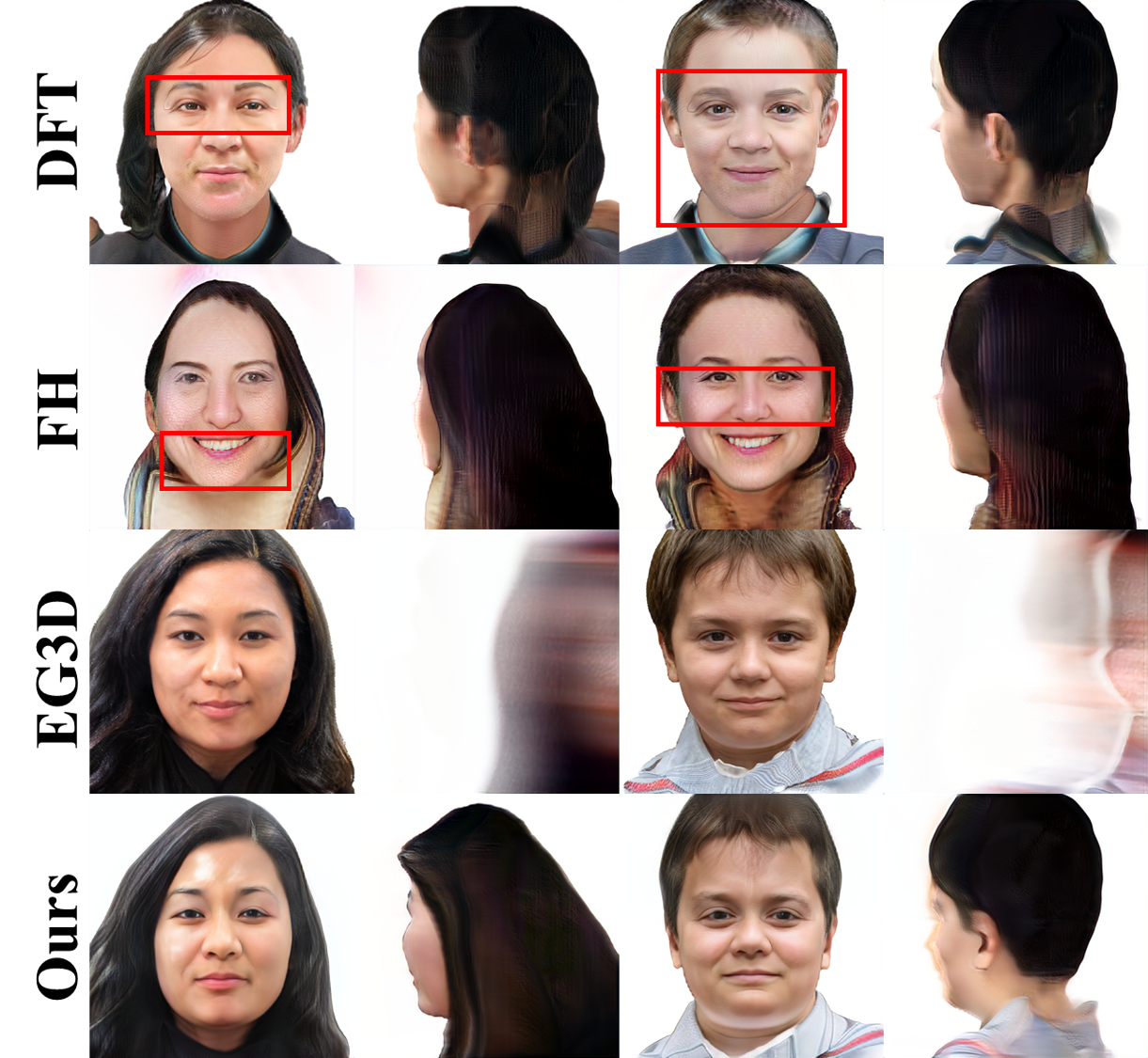}
	\caption{Qualitative comparisons between Head3D and baselines. There exhibit two perspectives of two identities.}
	\label{img:comparison}
 \vspace{-1em}
\end{figure}

As few works achieve full head generation trained by images, we compare our work with some designed baselines. 1) Directly fine-tune EG3D with FFHQ and H3DH, named \emph{DFT}. 2) Train model with FFHQ and H3DH from a fine-tuned head generator, named \emph{FH}. 3) Original pre-trained EG3D, to examine the quality of faces. The results are shown both qualitatively and quantitatively, where we use Fr$\acute{e}$chet Inception Distance (FID)~\cite{heusel2017gans} and Kernel Inception Distance (KID)~\cite{binkowski2018demystifying} for evaluation. Note that, we do not implement training from scratch on the H3DH, due to the fact that training with only 50 individuals results in insufficient diversity in the identity information.

\begin{table}[t]
	\centering
     \begin{tabular}{c|ccc}
			\hline
			Type & Baselines & FID $\downarrow$ & KID $\downarrow$  \\
			\hline
			Face &  EG3D~(Fine-tuned) & \textbf{6.82} & \textbf{0.36} \\
			 \hline
             \multirow{3}{*}{Head} &\emph{DFT} & 30.46 & 1.50\\
             &\emph{FH} &46.09 & 2.11\\
             &Ours & 11.34  & 0.61    \\
			\hline
	\end{tabular}
	\caption{Quantitative comparisons between our Head3D and baselines in FID and KID$\times100$, obtained in the resolution of $512^2$.} 
	\label{lab:comparison}
	\vspace{-1.5em}
\end{table}

\begin{figure}[t]
	\centering
	\includegraphics[width=\linewidth]{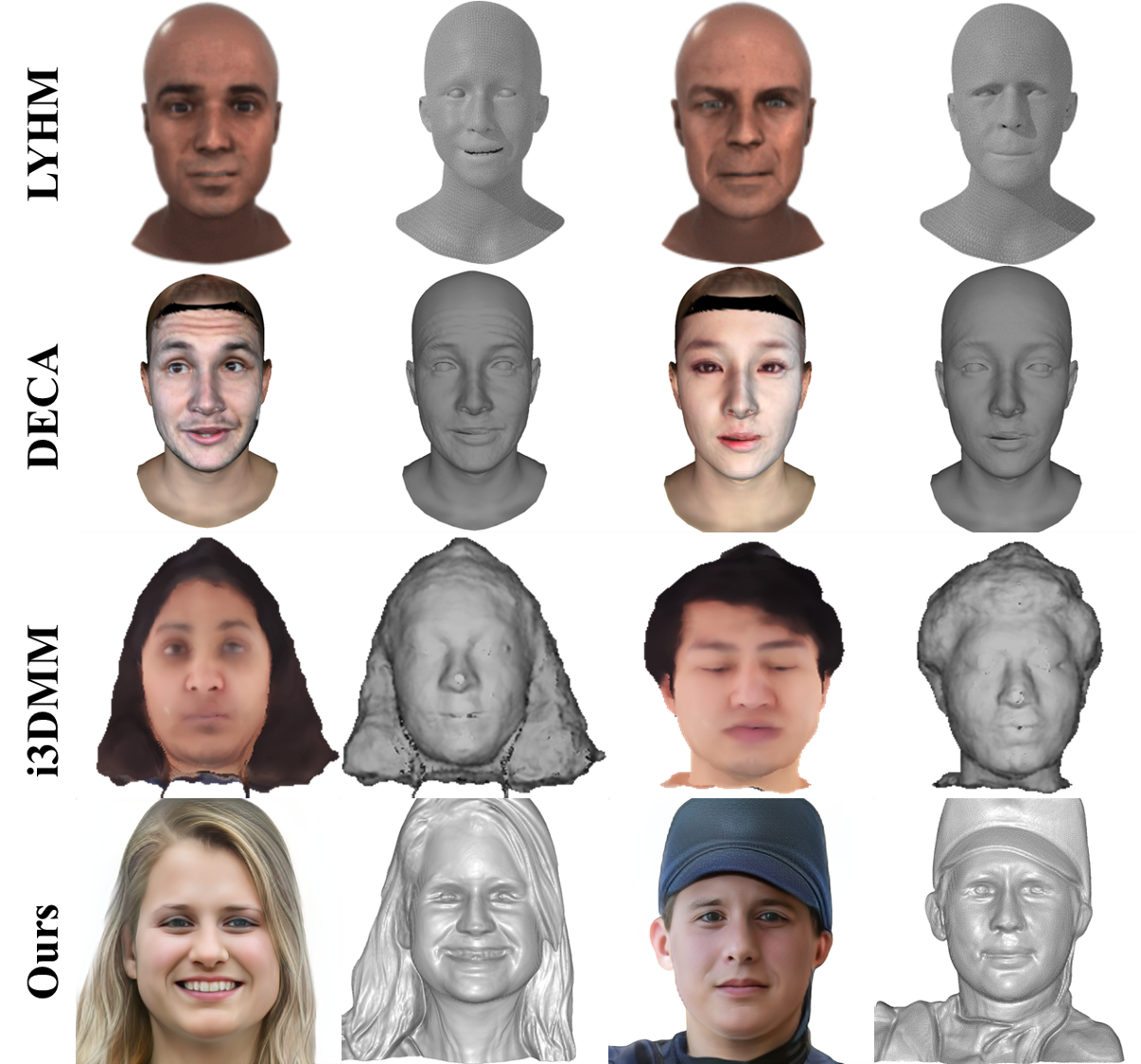}
	\caption{Qualitative comparisons among our Head3D and current generative head models. Results of LYHM~\cite{ploumpis2019combining,ploumpis2020towards}, DECA~\cite{feng2021learning} and i3DMM~\cite{yenamandra2021i3dmm} are obtained via their papers and released codes.}
	\label{i3dmm}
	\vspace{-2em}
\end{figure}

Figure~\ref{img:show} provides example results synthesized by our proposed method from various viewpoints and identities, demonstrating the generation of complete head geometry and high-quality renderings. Particularly, our method is able to generate accessory items such as hats and glasses, which are not present in the H3DH dataset. The visualized comparisons are shown in Figure~\ref{img:comparison}, where images in the same column are synthesized from the same latent code $z$. Although \emph{DFT} and \emph{FH} achieve full head generation, output consistency with origin EG3D is broken. Moreover, the model is trained by naively combined FFHQ and H3DH, resulting in low-quality results and model collapse. Similar observation can also be found from Table~\ref{lab:comparison} that both FID and KID become much worse in \emph{DFT} and \emph{FH}. Original EG3D performs the best numerically, while it can not generate full heads. Our work achieves identity representation consistent with the original EG3D through knowledge distillation and can fully represent human heads with good 3D consistency. 
Although our method is numerically inferior to the original EG3D, considering information loss in knowledge distillation and no back images corresponding to FFHQ faces, the decline in generating quality is acceptable. Especially, the generated heads maintain high-fidelity rendering results as EG3D qualitatively.
In summary, our H3DH achieves high-quality complete heads generation in both visual results and quantitative evaluation.

A qualitative comparisons are conducted among our Head3D and current generative head models, LYHM~\cite{ploumpis2019combining,ploumpis2020towards}, DECA~\cite{feng2021learning} and i3DMM~\cite{yenamandra2021i3dmm}, in terms of rendering results and geometry shapes, as depicted in Figure ~\ref{i3dmm}. Notably, these models are trained on the dataset consisting of a large number of 3D scans~(several hundreds or thousands), whereas our Head3D model only uses a multi-view image dataset containing 50 individuals. However, the results reveal that our model produces superior rendering quality and more detailed geometry than these methods. Additionally, explicit head models~\cite{ploumpis2019combining,ploumpis2020towards,feng2021learning} are not capable of representing hair, while the implicit model i3DMM~\cite{yenamandra2021i3dmm} can only represent hair in low quality. In contrast, our results show the ability to generate photo-realistic and diverse hair, including various accessories such as hats. Therefore, this comparison demonstrates that our algorithm can achieve high-quality and highly-detailed complete head generation with significantly fewer data.


\subsection{Ablation Study}

\begin{figure}[t]
\centering
    	\begin{minipage}[h]{1.0\linewidth}
    	\centering
    	\includegraphics[width=1.0\linewidth]{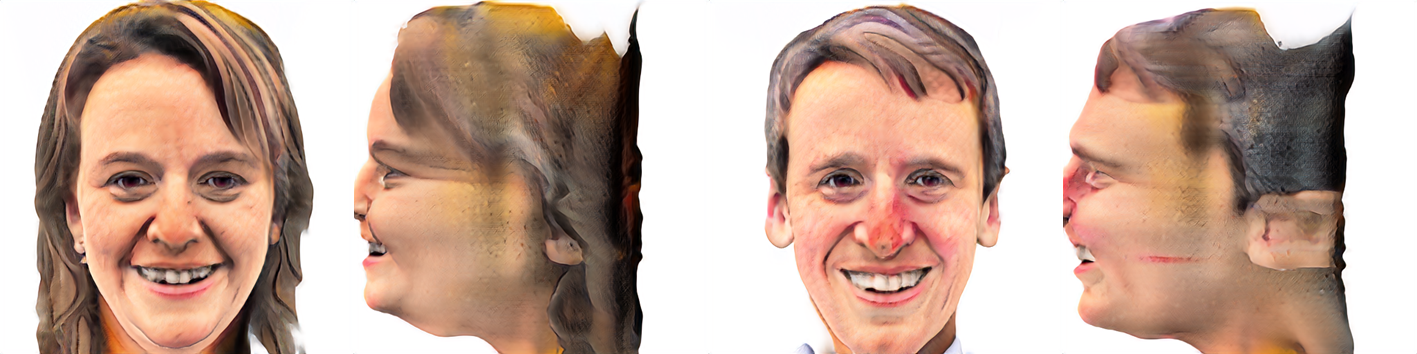}
    	\small (a)~Results of applying $f^t_{xy}$ on head generator without training.
    	\includegraphics[width=1.0\linewidth]{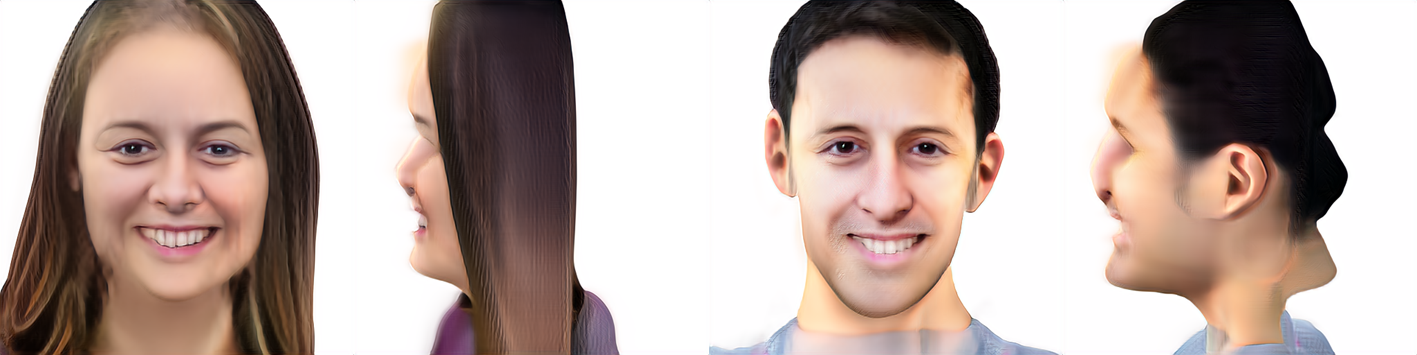}
    	\small (b)~Results of the model trained with total tri-plane distillation.
    	\includegraphics[width=1.0\linewidth]{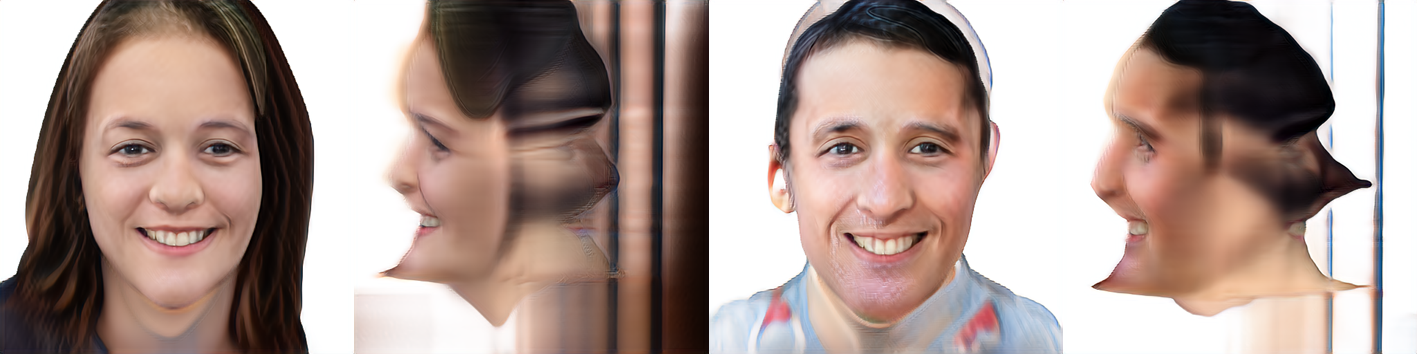}
    	\small (c)~Results of the model trained by single discriminator.
    	\includegraphics[width=1.0\linewidth]{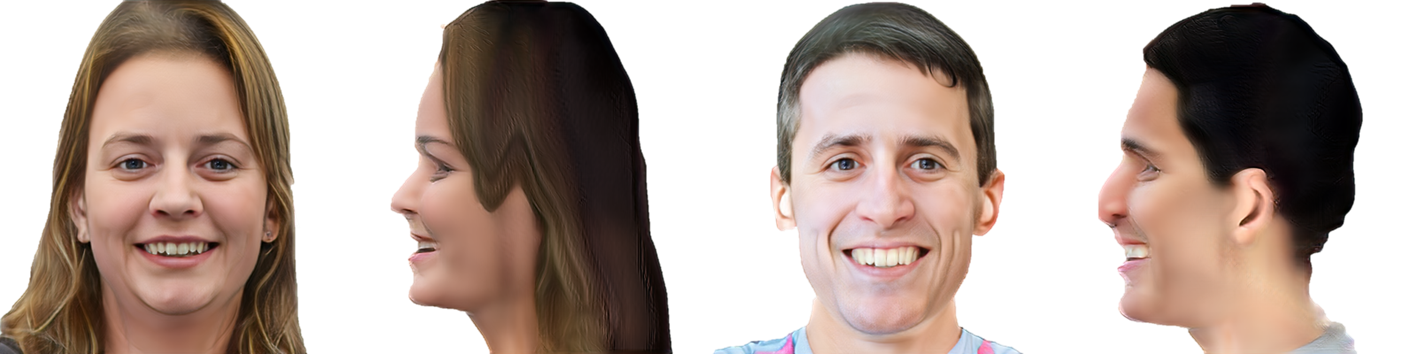}
    	\small (d)~Results of our Head3D.
	    \end{minipage}%
	\centering
	\caption{The illustration of ablation studies on tri-plane feature distillation and dual-discriminator. }
\vspace{-1em}
\label{img:triplane}
\end{figure}

\begin{table}[t]
	\centering
    \begin{tabular}{lccc}
			\hline
			& FID $\downarrow$ & KID $\downarrow$ & ID $\uparrow$ \\
			\hline
			W/O $L_{kd}$ & 44.76  & 3.07   & 0.10\\
			W/O $L_{gan_{front}}$ & 29.84 & 2.47& 0.45\\
			W/O $L_{rgb}\& L_{lpips}$ & 19.50 & 1.29& 0.45  \\
			Ours & \textbf{11.34}  & \textbf{0.61}  & \textbf{0.65} \\
			\hline
	\end{tabular}
	\caption{Quantitative results of ablation study on different loss functions, evaluated by FID, KID$\times100$ and ID consistency (ID).} 	
	\label{lab:ablation}
	\vspace{-1.5em}
\end{table}



\noindent \textbf{Effectiveness of Tri-plane Feature Distillation.} We investigates the effectiveness of our proposed tri-plane feature distillation method. We considered two settings: (a) directly applying $f^t_{xy}$ on the head generator without training, and (b) distilling the whole tri-plane features, as results shown in Figure~\ref{img:triplane} (a) and (b), respectively. It can be concluded in the first setting, although the identity information is preserved, the generated heads suffer from severe distortions and defective head shapes. In the second setting, distilling the whole tri-plane leads to incomplete head shapes. In contrast, our proposed method, as shown in Figure 3~(d), maintains identity consistency and achieves complete head generation. These results demonstrate the effectiveness of our tri-plane feature distillation approach.


\noindent \textbf{Effectiveness of Dual-discriminator.} We also conduct experiments to verify the effectiveness of our proposed dual-discriminator. As depicted in Figure~\ref{img:triplane} (c), the model trained with a single discriminator produces competitive results for face synthesis with Head3D. However, it fails to generate the back of the head resulting in a stitching of two faces. In summary, the comparison verifies the effectiveness of the proposed dual-discriminator.

\begin{figure}[t]
	\centering
	\includegraphics[width=\linewidth]{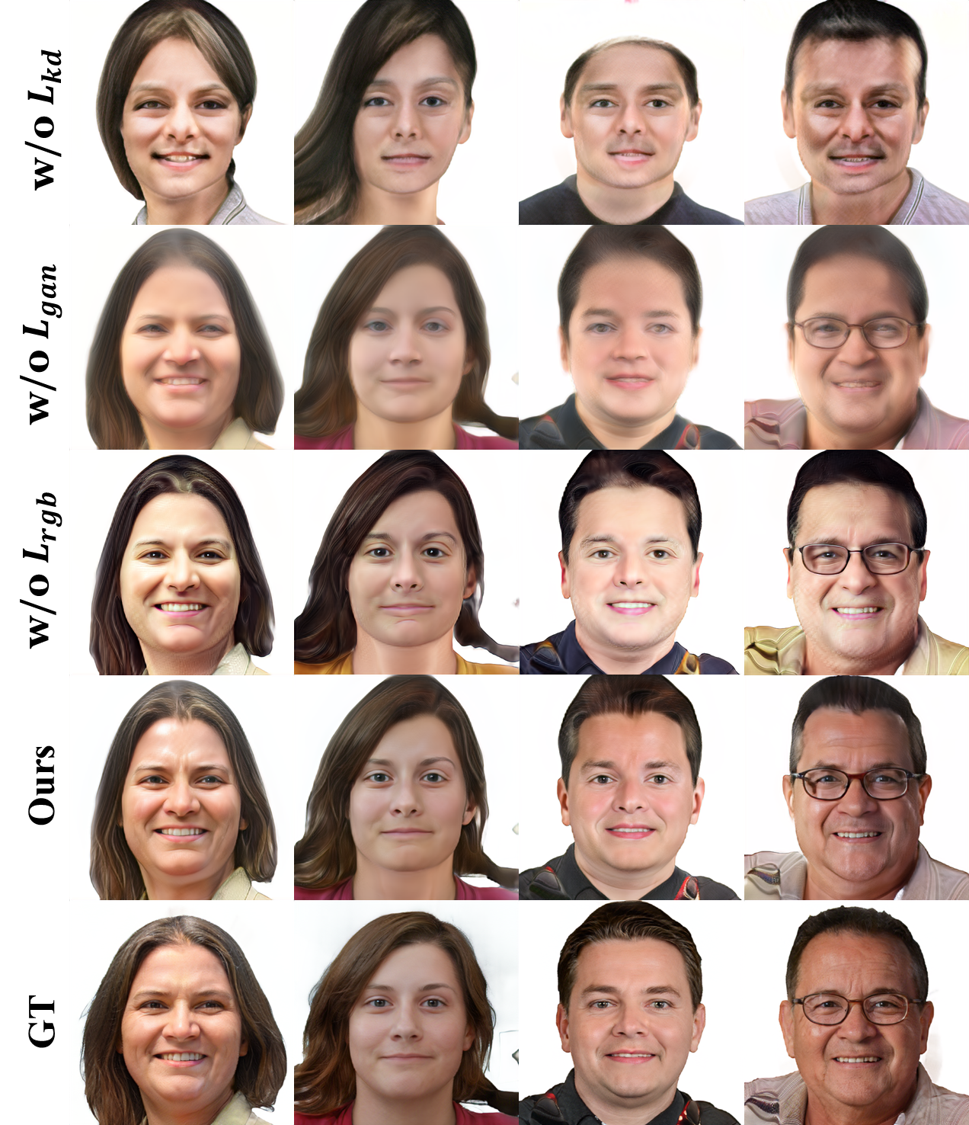}
	\caption{The ablation studies on the effectiveness of each loss function. All images in a column are generated from the same latent code. Note that, the ground-truth results are generated by the pre-trained face generator without background.}
	\label{img:ablations}
	\vspace{-1.5em}
\end{figure}

\noindent \textbf{Effectiveness of Loss Functions.}
We conduct an ablation study to verify the effectiveness of each losses. Specifically, we remove each loss function individually and maintain other settings the same. The results are evaluated both qualitatively and quantitatively. The quantitative results including FID, KID and Identity Distance (ID) to the origin EG3D, calculated by Arcface~\cite{deng2019arcface} among 10000 individuals, are presented in Table~\ref{lab:ablation}. The results show that removing any loss functions results in a significant increase in FID and KID scores. Moreover, the lowest ID score is obtained when knowledge distillation is removed, which highlights the crucial role of tri-plane distillation in transferring identity information.

Figure~\ref{img:ablations} presents example results of different settings to further evaluate the effectiveness of each loss function. Although most of the textures are preserved without $L_{gan_{front}}$, the photo-realism is lost, which indicates the importance of adversarial training for maintaining image quality. While lacking $L_{kd}$ causes failure in identity transmission. $L_{rgb}$ and $L_{lpips}$ are also crucial in preserving detailed texture and ensuring texture consistency. 
Overall, the combination of these loss functions achieves the best results both quantitatively and qualitatively.

 \begin{figure}[t]
	\centering
	\includegraphics[width=\linewidth]{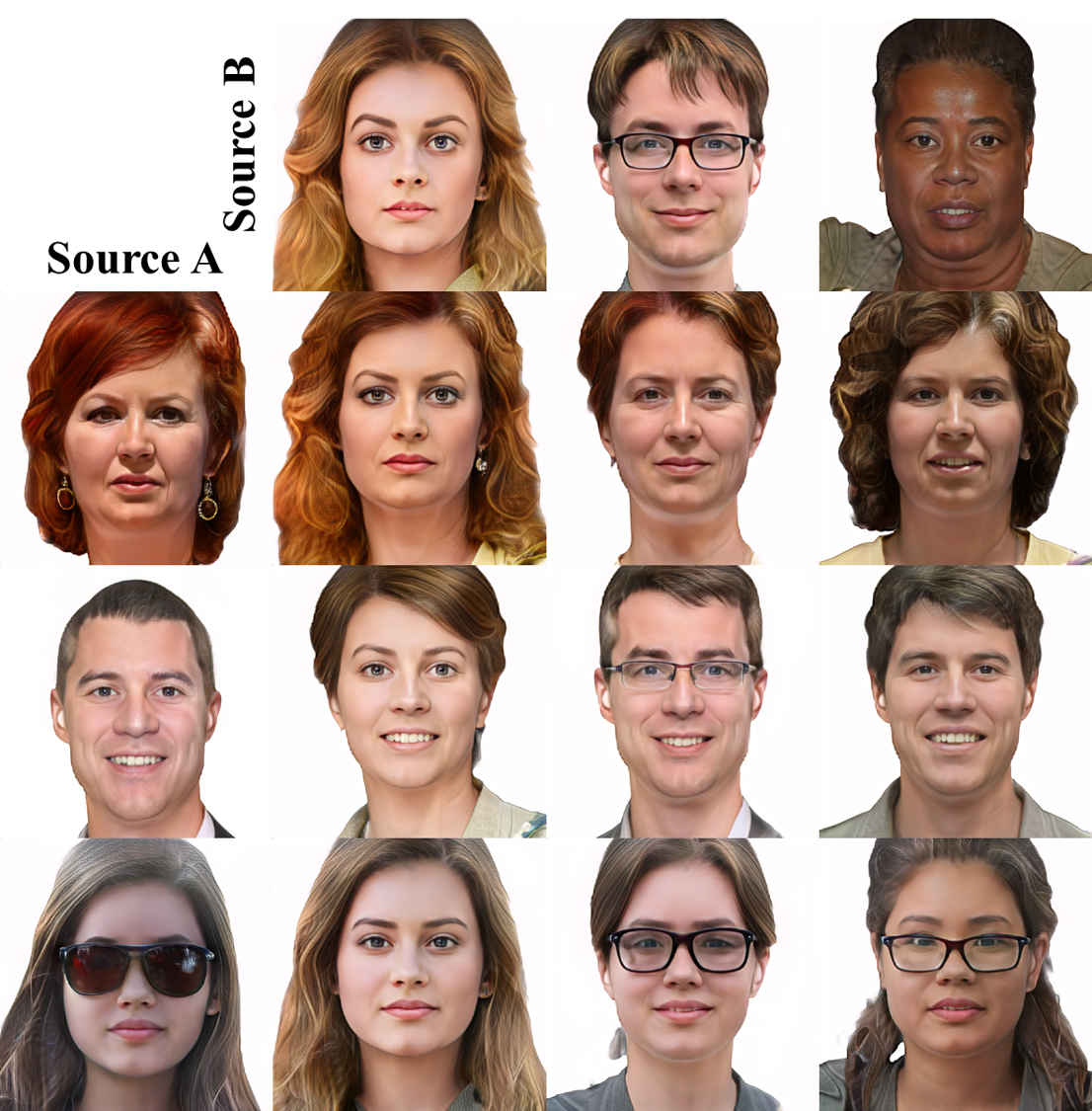}
	\caption{Interpolation results of our proposed Head3D model.}
	\label{img:stylemix}
	\vspace{-1.5em}
\end{figure}


\noindent\textbf{Analysis of Linearity in Latent Space.}
Analogous to StyleGAN-based generators \cite{karras2019style,Karras_2020_CVPR}, style codes $w$ can be linearly interpolated to achieve image manipulations. Figure~\ref{img:stylemix} demonstrate that our model remains linear separability in the latent space after knowledge distillation.
\section{Conclusion}

This paper presents Head3D, a method for generating complete heads trained with limited data. We first revisit the EG3D framework and emphasize the importance of tri-plane as a semantic information carrier. Through experiments, we demonstrate that tri-plane decoupling is achievable, with identity information controlled by the $xy$-plane. We then propose a tri-plane feature distillation approach and a dual-discriminator method for training head generators. Extensive experiments confirm the effectiveness of our proposed method. We hope that our work will inspire further researches in generating diverse and high-quality 3D complete heads from limited and uncorrelated images.




{\small
\bibliographystyle{ieee_fullname}
\bibliography{eg3dkd}
}

\end{document}